\documentclass[sigconf]{acmart}
\AtBeginDocument{%
  }

\copyrightyear{2026}
\acmYear{2026}
\acmConference[ICMI '26]{INTERNATIONAL CONFERENCE ON MULTIMODAL INTERACTION}{October 05--09, 2026}{Napoli, Italy}
\acmBooktitle{INTERNATIONAL CONFERENCE ON MULTIMODAL INTERACTION (ICMI '26), October 05--09, 2026, Napoli, Italy}
\acmDOI{10.1145/3776574.3832485}
\acmISBN{979-8-4007-2318-6/2026/10}

\acmSubmissionID{1002}

\usepackage{booktabs}
\usepackage{tikz}
\usetikzlibrary{positioning,arrows.meta,shapes.geometric,fit,backgrounds,calc}

\begin{document}

\title{Mind the Student: Behavioral and Contextual Cues for Automated Engagement Prediction in Online Learning}

\author{Alperen Kantarcı}

\orcid{0000-0002-4080-5538}
\correspondingauthor
\authornotemark[1]
\affiliation{%
  \institution{Institute of Computer Science}
\institution{Goethe University Frankfurt}
  \city{Frankfurt am Main}
  \country{Germany}
}
\email{kantarci@em.uni-frankfurt.de}

\author{Visvanathan Ramesh}
\orcid{0000-0002-8842-905X}
\affiliation{%
 \institution{Institute of Computer Science}
 \institution{Goethe University Frankfurt}
  \institution{The Hessian Center for Artificial Intelligence}
  \city{Frankfurt am Main}
  \country{Germany}}
\email{vramesh@em.uni-frankfurt.de}

\author{Gemma Roig}
\orcid{0000-0002-6439-8076}
\affiliation{%
 \institution{Institute of Computer Science}
 \institution{Goethe University Frankfurt}
 \institution{The Hessian Center for Artificial Intelligence}
  \city{Frankfurt am Main}
  \country{Germany}}
\email{roignoguera@em.uni-frankfurt.de}

\renewcommand{\shortauthors}{Kantarcı et al.}

\begin{abstract}
The prediction of student engagement from the online tutoring videos is difficult because engagement is a multidimensional construct comprising distinct behavioral, emotional, and cognitive states. A reliable prediction requires bringing together different types of behavioral signals as well as expressive cues. Through our analysis of the CASED dataset, it is clear that engagement prediction gets even harder due to the high inter-person variability as well as the subjectivity of the engagement annotation. To tackle these challenges, we develop a multimodal framework that integrates the implicit spatiotemporal features extracted from pretrained video, audio, and image encoders along with structured behavioral modalities like head pose, gaze, facial action units, emotion, and wavelet-based audio features. We integrate these modalities via a Perceiver IO latent bottleneck. Moreover, student and instructor personalities are modeled as variational posteriors over learnable embeddings to enable partial pooling across participants. We employ evidential regression and spectral-normalized Gaussian process classification heads for uncertainty-aware prediction to further improve robustness and calibration. Benchmark on the CASED challenge test set shows that all participating methods converge near random-chance performance, revealing the difficulty of the dataset. In this highly ambiguous regime, our framework achieves competitive performance while uniquely offering well-calibrated uncertainty metrics, demonstrating that reliable risk-quantification is an essential prerequisite for deploying engagement models in real-world educational tools.
\end{abstract}



\setcopyright{none}
\maketitle

\section{Introduction and Background}
Student engagement is widely recognized as a fundamental prerequisite for effective learning and academic success~\cite{bergdahl2024unpacking,engagement_zimmer}. From the perspective of educational psychology, engagement is a multidimensional construct with behavioral, emotional, cognitive, and social dimensions that collectively reflect a learner's involvement in the educational process. High levels of engagement have been associated with improved learning outcomes, higher knowledge retention, enhanced motivation, and reduced dropout rates~\cite{fredricks2004school}. Therefore, understanding and analyzing student engagement is an important objective for educational researchers.

A significant challenge is that engagement indicators are distributed across different  modalities. A student's face and posture can reflect their attention, or an instructor's behavior can influence responsiveness. The  screen content can affect cognitive load and audio can convey speech-based information~\cite{screencontent}. Moreover, these signals are not  equally informative in every instance. For example, gaze direction may become unreliable when the face is partially obscured, or emotion predictions may become unstable at low resolutions. This necessitates the development of architectures capable of integrating diverse modalities  while remaining robust against varying signal quality.

\begin{figure*}[t]
    \centering
    \includegraphics[width=0.8\linewidth]{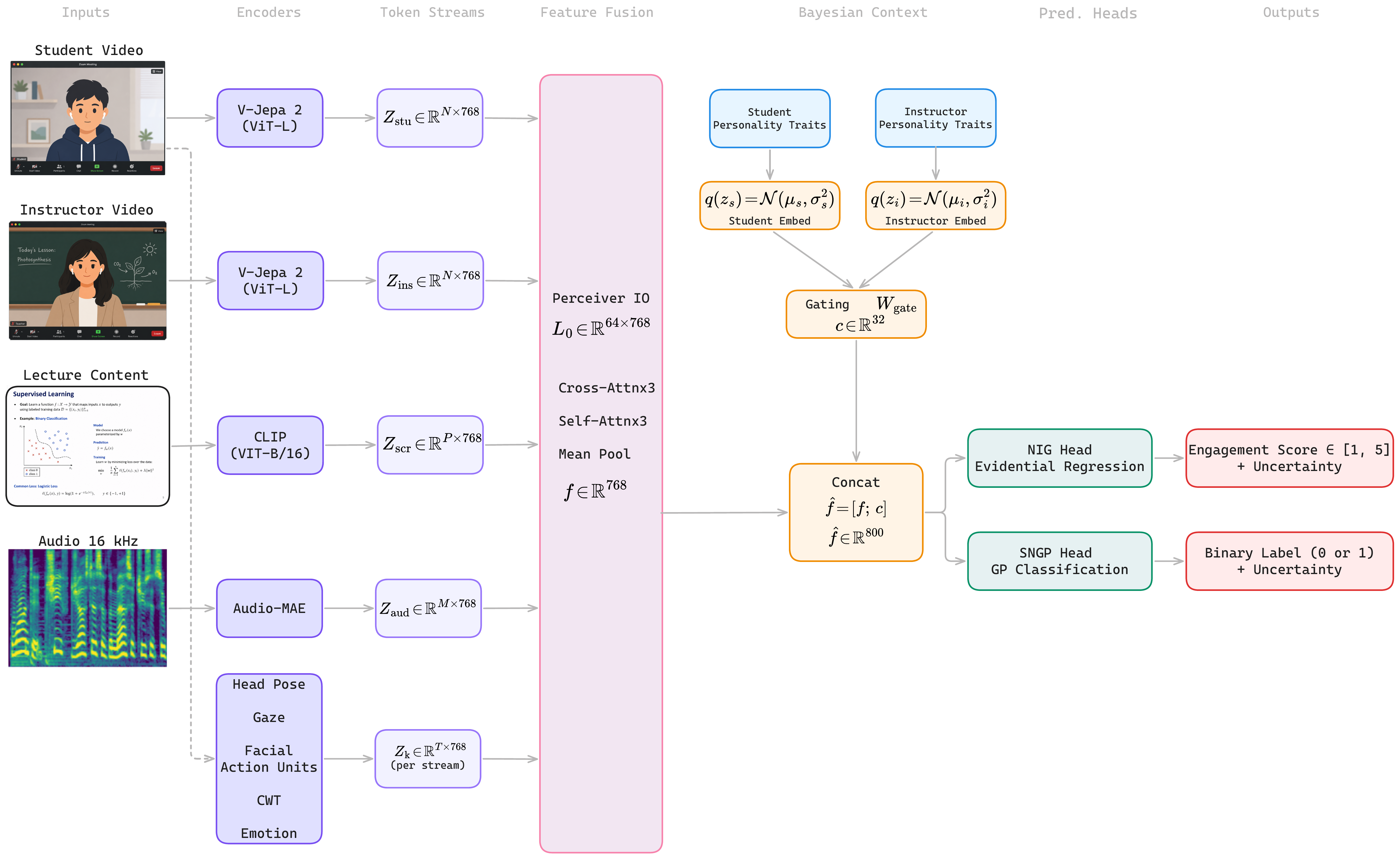} 
    \caption{Overview of the proposed multimodal engagement framework. Multi-source inputs (student/instructor video, slides, audio, and behavioral cues) are encoded via specialized models and fused using a Perceiver IO block to form the core feature representation (d=768) This is combined with a personality-driven Bayesian Context embedding  via a gating mechanism. The final concatenated representation (d=32) feeds into an Evidential Regression head and a GP Classification head to simultaneously predict engagement levels and explicit epistemic uncertainty.}
    \label{fig:model_overview}
\end{figure*}


The CASED~\cite{cased,cased2} challenge reflects this difficulty. Predicting student engagement from short clips of online tutoring sessions is complicated by several factors: limited training data, substantial inter-person variability, the inherent subjective nature of engagement annotation~\cite{annotation,inconsistency}. In these settings, predicting performance can be near-random or weakly above-baseline results. This can indicate not only model limitations but also dataset ambiguity, label noise, or insufficient signal in the available observations. Therefore, methods for this task should be evaluated not only by predictive accuracy, but also by how well they handle uncertainty, exploit multimodal structure, and generalize to unseen students.

To this end, we present a multimodal engagement prediction framework that combines behavioral cues using large pretrained encoders, efficient multimodal fusion, participant-level Bayesian context modeling, and uncertainty-aware prediction heads. Particulary, we extract explicit behavioral features including head pose, gaze, facial action units, emotion estimates and more generic representations from student video, instructor video, screen content and audio. These various features are fused using Perceiver IO~\cite{perceiverio} which is an modality-agnostic asymmetric attention mechanism. We also introduce a hierarchical Bayesian context layer over student and instructor embeddings with partial pooling toward a shared prior to model person-specific effects without encouraging identity memorization.
\subsection{Proposed method}
Given a video clip \(V\), we jointly predict a continuous engagement score \(y \in [1,5]\) and a binary label \(\ell \in \{0,1\}\). Each clip is associated with a student identity \(s\), an instructor identity \(i\), and optional personality vectors \(t_s,t_i \in \mathbb{R}^{10}\). Evaluation uses a student-independent split which means training and testing sets have different students.


Each clip is decomposed into student video, instructor video, screen content, audio, and explicit behavioral features. For \(1280\times1280\) composite videos, we crop fixed student, instructor, and screen regions, resize them to \(224\times224\), and uniformly sample \(T=64\) frames for student and instructor streams. Audio is resampled to 16 kHz mono and converted to a Kaldi filterbank~\cite{kaldiaudio} spectrogram of shape \((1024,128)\).

We use pretrained encoders for the main modalities: V-JEPA 2~\cite{vjepa2} for student and instructor video, CLIP~\cite{clip} for the  screen content, and AudioMAE~\cite{audiomae} for the spectrogram. Overall architecture visualization can be seen in Figure~\ref{fig:model_overview}. Their outputs are projected to a shared dimension, d=768. We utilize a single Linear Layer for each modality, followed by Layer Normalization to stabilize the inputs before they enter the Perceiver IO bottleneck.
$
Z_{\text{student}}, Z_{\text{instructor}} \in \mathbb{R}^{N\times768},\quad
Z_{\text{screen}} \in \mathbb{R}^{P\times768},\quad
Z_{\text{audio}} \in \mathbb{R}^{M\times768}.
$

We additionally extract five explicit feature streams at \(T=64\): head pose, CWT head dynamics, facial action units, gaze/head angles, and emotion prediction probabilities. Each feature sequence \(F_k \in \mathbb{R}^{T\times d_k}\) is linearly projected as
$Z_k = F_k W_k \in \mathbb{R}^{T\times768}$.


All token streams are fused with Perceiver IO~\cite{perceiverio}. Let \(\mathcal{M}\) be the set of modalities and \(e_{c_k}\) the learned embedding for modality \(k\). The input token set is
\begin{equation}
X = \mathrm{concat}\left[ Z_k + e_{c_k} \mid k \in \mathcal{M} \right]
\in \mathbb{R}^{T_{\mathrm{total}}\times768}.
\end{equation}

A latent array \(L_0 \in \mathbb{R}^{Q\times768}\) with \(Q=64\) queries is updated for \(L=3\) layers via cross-attention to \(X\) and latent self-attention.
The fused representation is obtained by mean pooling:
\begin{equation}
f = \mathrm{mean}(\mathrm{LN}(L_L)) \in \mathbb{R}^{768}.
\end{equation}

\paragraph{Bayesian Participant Context:}

To model stable person-specific effects, we assign each student \(s\) and instructor \(i\) a variational embedding:
\begin{equation}
q(z_s)=\mathcal{N}(\mu_s,\mathrm{diag}(\exp(\sigma_s^2))),\quad
q(z_i)=\mathcal{N}(\mu_i,\mathrm{diag}(\exp(\sigma_i^2))).
\end{equation}
Embeddings are sampled using the reparameterization trick at the training time. Posterior means are used, and unseen students receive a learned prior mean \(\mu_{\text{prior}}\) during the inference.

When personality metadata are available, they are added through
\begin{equation}
\tilde z_s = z_s + m\,\tanh(W_{\text{trait}} t_s),
\end{equation}
where \(m\) is a binary mask dropped during training and set to zero at test time. Student and instructor context are combined by
$c = \sigma(W_{\text{gate}}[\tilde z_s;\tilde z_i]) \in \mathbb{R}^{32}$, and concatenated with the fused representation:
$\hat f = [f;c] \in \mathbb{R}^{800}.$

A KL penalty regularizes participant posteriors toward a shared prior:
\begin{equation}
\mathcal{L}_{\text{KL}}
=
\frac{1}{N}\sum_s
D_{\mathrm{KL}}
\!\left(
\mathcal{N}(\mu_s,\sigma_s^2)\,\|\,\mathcal{N}(\mu_{\text{prior}},I)
\right).
\end{equation}

\paragraph{Uncertainty-Aware Prediction Heads:}

For regression, we use an evidential Normal-Inverse-Gamma (NIG) head \cite{amini2020deep}, which predicts $(\gamma,\nu,\alpha,\beta)$, 
where \(\gamma\) is the engagement estimate and the remaining parameters define predictive uncertainty. The corresponding aleatoric and epistemic uncertainties are
$\beta / (\alpha-1)$ and $\beta/(\nu(\alpha-1)).$

For classification, we use a spectral-normalized neural Gaussian process (SNGP) head~\cite{sngp}, with predictive probability
\begin{equation}
p(\ell=1\mid \hat f)=
\sigma\!\left(
\frac{w^\top \hat f}{\sqrt{1+\pi\kappa^2/8}}
\right),
\end{equation}
where \(\kappa^2\) is the predictive variance.

\paragraph{Training Objective:}

The total loss is
\begin{equation}
\mathcal{L}
=
\frac{
\left(
\lambda_{\text{NIG}}\mathcal{L}_{\text{NIG}}
+
\lambda_{\text{CCC}}\mathcal{L}_{\text{CCC}}
\right)}{e^{\sigma_r}}
+\sigma_r
+
\frac{
\left(
\lambda_{\text{cls}}\mathcal{L}_{\text{cls}}
+
\lambda_{\text{ord}}\mathcal{L}_{\text{ord}}
\right)}{e^{\sigma_c}}
+\sigma_c
+\mathcal{L}_{\text{KL}},
\end{equation}
where \(\sigma_r,\sigma_c\) are learnable task-uncertainty weights \cite{kendall2018multi}. We use NIG loss and CCC loss for regression, weighted cross-entropy for classification, and an auxiliary ordinal consistency loss coupling regression and classification outputs.

\paragraph{Optimization and Ensemble:}

We train with AdamW~\cite{DBLP:conf/iclr/LoshchilovH19}, weight decay \(5\times10^{-2}\), cosine decay, and 5\% warmup. Backbone encoders use learning rate \(10^{-5}\), while newly initialized layers use \(5\times10^{-4}\). Training is staged: encoders are frozen for epochs 1--20 and unfrozen until the convergance. We additionally use random temporal sampling, metadata dropout, and KL (Kullback-Leibler) annealing. For our final predictions we use four different training checkpoints of the same model as ensemble of networks and do a majority voting on the predictions.

\section{Experiments and Dataset Analysis}
We use DaiSEE~\cite{daisee} and Aff-Wild2~\cite{abaw1,abaw2,abaw3} datasets for pretraining and CASED~\cite{cased,cased2} dataset for fine-tuning. We evaluate under a strict student-independent protocol. For our trainings and validation experiments, we use 5-fold cross-validation with student-level fold assignment. Final leaderboard performances are reported from CASED test set. We report F1 Macro, F1 Weighted, MCC, RMSE, MSE, MAE, $R^2$, Pearson Correlation and Concordance Correlation Coefficient (CCC) depending on Classification and regression tasks. 
\subsection{Dataset analysis}
The binary label distribution is highly skewed: approximately 69\% of clips are labeled engaged (label 0) and 31\% not-engaged (label 1). The continuous engagement scores are similarly compressed, with mean 3.7 and median 4.0, producing a long lower tail. A majority-class classifier therefore achieves approximately 40–45\% macro F1 without learning any discriminative features.

\paragraph{Student Analysis:} Individual students differ substantially in their mean engagement level and within-session variability. Several students are labeled engaged in over 90\% of their clips while others fall below 40\%. This hints that a large fraction of the label variance is explained by between-student differences rather than within-clip behavioral dynamics, which directly limits what any clip-level visual model can learn.

\paragraph{Instructor Analysis:} Per-instructor engagement rates do not vary systematically. All three instructors have 3.37 mean engagement with similar standard deviation. Around 68\% of clips are labeled as engaged for all instructors. Furthermore, all students are paired with exactly one instructor across all sessions, making student and instructor identity perfectly collinear. 

\section{Results}

Table \ref{tab:classification} and Table \ref{tab:regression} summarize the official challenge leaderboard for the classification and regression tracks. Our method performs competitively among participating teams. However, the margin between the top methods is small. Best-performing method obtained a F1 Macro of 0.52,  suggesting that all approaches face with similar limitations imposed by the dataset and the task. Other metrics also show very similar performances.
In the regression task, our approach achieved a matching or closely matching the best-performing methods. However, all submissions obtained near-zero or negative values for $R^2$, Pearson correlation, and CCC, indicating that none of the evaluated methods was able to reliably capture the underlying engagement signal. Collectively, the leaderboard suggests that the proposed framework performs competitively while highlighting the intrinsic difficulty of automatic engagement prediction on this benchmark.
\begin{table}[h]
\centering
\caption{Classification performance of participating methods on the CASED challenge test set. }
\label{tab:classification}
\resizebox{\columnwidth}{!}{%
\begin{tabular}{lrrrrrr}
\toprule
   Participant &   F1 (Macro) &  F1 (Weighted) &  Precision &  Recall &  MCC \\
\midrule
        saurabhh &         0.52 &           0.60 &       0.52 &    0.52 & 0.04 \\
         mohitvu &         0.52 &           0.61 &       0.52 &    0.52 & 0.04 \\
          adim66 &         0.51 &           0.58 &       0.51 &    0.52 & 0.03 \\
 \textbf{Ours} &         0.51 &           0.60 &       0.51 &    0.51 & 0.02 \\
       priscalab &         0.51 &           0.60 &       0.51 &    0.51 & 0.01 \\
         caymann &         0.50 &           0.58 &       0.50 &    0.50 & 0.01 \\
    KvochurHegel &         0.42 &           0.59 &       0.35 &    0.50 & 0.00 \\
\bottomrule
\end{tabular}
}
\end{table}

\begin{table}[h]
\centering
\caption{Regression performances on the CASED challenge test set.}
\label{tab:regression}
\resizebox{\columnwidth}{!}{
\begin{tabular}{lrrrrrrr}
\toprule
      Participant  &  RMSE &  MSE &  MAE &  $R^2$ &  Pearson &   CCC \\
\midrule
        saurabhh &     0.68 & 0.46 & 0.57 &  -0.01 &    -0.00 & -0.00 \\
         mohitvu &     0.78 & 0.61 & 0.63 &  -0.33 &    -0.00 & -0.00 \\
 \textbf{Ours} &     0.68 & 0.46 & 0.57 &  -0.00 &    -0.02 & -0.00 \\
       priscalab &     0.68 & 0.47 & 0.57 &  -0.02 &     0.01 &  0.00 \\
         caymann &     0.68 & 0.47 & 0.57 &  -0.01 &     0.03 &  0.01 \\
    KvochurHegel &     0.68 & 0.47 & 0.57 &  -0.01 &    -0.03 & -0.00 \\
\bottomrule
\end{tabular}
}
\end{table}

\paragraph{Multimodal and Component Ablation}
We evaluate the impact of different modality combinations and architectural components on engagement prediction. As shown in Table~\ref{tab:ablation}, the inclusion of all three modalities—Student, Instructor, and Content (S+I+C)—yields the highest performance, achieving a CCC of 0.018 and an F1-macro score of 0.498. This confirms that student engagement is heavily contextual and benefits from integrating student behavior alongside instructor and screen dynamics.
In the component ablation Table~\ref{tab:ablation}, the Full Model outperforms or remains highly competitive with alternative configurations. The base transformer baseline achieves lower RMSE and MAE. The integration of the Bayesian Context layer, NIG and SNGP heads optimizes macro-level classification and alignment metrics, yielding the top F1-macro and a strong CCC scores.
\begin{table}[h]
\centering
\caption{Ablation study. (a) Modality contribution using the full model (BayesianContext + NIG + SNGP). (b) Component contribution on the three-modality input (S=student video, I=instructor video, C=content). Metrics are averaged over 5 student-grouped cross-validation folds.}
\label{tab:ablation}
\resizebox{\columnwidth}{!}{
\begin{tabular}{llccccc}
\toprule
Model & Modalities & CCC $\uparrow$ & RMSE $\downarrow$ & MAE $\downarrow$ & F1-macro $\uparrow$ & MCC $\uparrow$ \\
\midrule
\multicolumn{7}{l}{\textit{(a) Modality ablation — Full model}}\\
\midrule
Full (S) & Student & -0.016 {\scriptsize$\pm$0.027} & 1.048 {\scriptsize$\pm$0.060} & 0.848 {\scriptsize$\pm$0.046} & 0.486 {\scriptsize$\pm$0.015} & -0.002 {\scriptsize$\pm$0.029} \\
Full (S+I) & Student + Instructor & -0.021 {\scriptsize$\pm$0.016} & 1.020 {\scriptsize$\pm$0.119} & 0.832 {\scriptsize$\pm$0.111} & 0.484 {\scriptsize$\pm$0.022} & -0.011 {\scriptsize$\pm$0.016} \\
Full (S+I+C) & Student + Instructor + Content & 0.018 {\scriptsize$\pm$0.029} & 0.943 {\scriptsize$\pm$0.038} & 0.755 {\scriptsize$\pm$0.040} & 0.498 {\scriptsize$\pm$0.019} & 0.002 {\scriptsize$\pm$0.034} \\
\midrule
\multicolumn{7}{l}{\textit{(b) Component ablation — Full modality set (S+I+C)}}\\
\midrule
Base Transformer & S+I+C & 0.015 {\scriptsize$\pm$0.028} & 0.821 {\scriptsize$\pm$0.022} & 0.671 {\scriptsize$\pm$0.026} & 0.497 {\scriptsize$\pm$0.013} & 0.012 {\scriptsize$\pm$0.019} \\
+ NIG + SNGP & S+I+C & 0.008 {\scriptsize$\pm$0.018} & 0.935 {\scriptsize$\pm$0.098} & 0.759 {\scriptsize$\pm$0.093} & 0.491 {\scriptsize$\pm$0.017} & 0.002 {\scriptsize$\pm$0.016} \\
+ Bayesian Context & S+I+C & 0.019 {\scriptsize$\pm$0.021} & 0.848 {\scriptsize$\pm$0.047} & 0.691 {\scriptsize$\pm$0.041} & 0.492 {\scriptsize$\pm$0.018} & 0.017 {\scriptsize$\pm$0.007} \\
\textbf{Full Model (Ours)} & \textbf{S+I+C} & \textbf{0.018 {\scriptsize$\pm$0.029}} & \textbf{0.943 {\scriptsize$\pm$0.038}} & \textbf{0.755 {\scriptsize$\pm$0.040}} & \textbf{0.498 {\scriptsize$\pm$0.019}} & \textbf{0.002 {\scriptsize$\pm$0.034}} \\
\bottomrule
\end{tabular}
}
\end{table}

\begin{table}[b]
\centering
\caption{NIG uncertainty decomposition by engagement level. Samples are grouped into five equal-width bins on the 1–5 continuous engagement scale.}
\label{tab:nig_uncertainty}
\resizebox{\columnwidth}{!}{
\begin{tabular}{lccccc}
\toprule
Engagement Level & N & Mean Pred & Aleatoric $\downarrow$ & Epistemic $\downarrow$ & Total \\
\midrule
1.0–1.8 & 41 & 3.45 & 42.0879 & 3323.3606 & 3365.4482 \\
1.9–2.6 & 657 & 3.37 & 33.1598 & 2623.2993 & 2656.4590 \\
2.7–3.4 & 1447 & 3.36 & 34.5978 & 2726.2368 & 2760.8345 \\
3.5–4.2 & 2398 & 3.38 & 34.7966 & 2802.9478 & 2837.7444 \\
4.3–5.0 & 435 & 3.42 & 36.4595 & 3008.4961 & 3044.9551 \\
\midrule
\multicolumn{6}{l}{\footnotesize Pearson $r$ (engagement vs aleatoric): 0.016; vs epistemic: 0.025.} \\

\bottomrule
\end{tabular}
}
\end{table}

\paragraph{Uncertainty Decomposition}
Table \ref{tab:nig_uncertainty} analyzes the uncertainty captured by the NIG regression head across different ground-truth engagement levels. Total uncertainty is heavily dominated by epistemic uncertainty across all bins. 
Pearson correlation coefficients show negligible linear relationships between raw engagement scores and both aleatoric and epistemic uncertainties, indicating that model confidence is driven by factors other than the absolute magnitude of the engagement score itself.
\begin{table}[h]
\centering
\caption{Bayesian context layer posterior statistics. The KL divergence from the population prior and the posterior mean norm are shown. 
}
\label{tab:bayesian_context}
\resizebox{\columnwidth}{!}{
\begin{tabular}{llccc}
\toprule
Entity & Name & KL from Prior $\downarrow$ & $\|\mu\|_2$ & N Clips \\
\midrule
All students (mean) & — & 14.3253 & 0.0273 & 116 \\
All instructors (mean) & — & 13.1287 & 0.0167 & 1659 \\
\midrule
\multicolumn{5}{l}{\textit{Top-3 students by KL (most personalized)}}\\
Student & laurencedu & 15.7200 & 0.0257 & 26 \\
Student & pierreyoussef & 15.4660 & 0.0795 & 50 \\
Student & richard & 15.2584 & 0.0311 & 55 \\
\midrule
\multicolumn{5}{l}{\textit{Bottom-3 students by KL (least personalized)}}\\
Student & administrator & 13.4648 & 0.0191 & 116 \\
Student & sophie & 13.3617 & 0.0205 & 129 \\
Student & akhat & 12.6692 & 0.0165 & 246 \\
\midrule
\multicolumn{5}{l}{\textit{Instructors}}\\
Instructor & Nigel Lu & 11.7067 & 0.0149 & 706 \\
Instructor & Pierre & 11.4088 & 0.0122 & 1454 \\
Instructor & Catherine & 11.2344 & 0.0111 & 2818 \\
\bottomrule
\end{tabular}
}
\end{table}

\paragraph{Bayesian Context Layer Personalization}
We evaluate the behavioral shifts in the learned identity embeddings via the KL divergence from the population prior.	
Table \ref{tab:bayesian_context} shows student identities experience a higher deviation from the population prior (mean KL=14.32) compared to instructor identities (mean KL=13.12), indicating stronger personalization for individual learners.
Moreover, statistical analysis reveals a strong, significant negative correlation between a student's clip count and their KL divergence from the prior (r=-0.748, p=0.001). This suggests that the model applies aggressive, highly individualized posterior shifts to sparse data. Conversely, students with abundant training clips converge closer to a well-regularized population norm.

\section{Limitations and Discussion}
The proposed method has several limitations due to both data and architectural choices. Most of evaluated configurations on test set yield validation CCC below 0.02 and F1-macro not more than 0.52, indicating models do not consistently outperform a constant mean predictor. There are several possible reasons for this. First, the not-engaged class drives the dominant failure mode. Clips near the Likert midpoint occupy an annotation boundary where small annotator perturbations flip the binary label. Secondly, most of representations remain identity-discriminative, causing the model to fit student appearance rather than engagement dynamics. The population prior in the Bayesian context layer is too weakly constrained to compensate. 
Finally, each clip receive a single label while Perceiver IO mean-pools over 64 frames, suppressing within-clip fluctuations. Finer-grained temporal supervision would be helpful for detecting small engagement cues.
\section{Conclusion}

We presented a multimodal engagement prediction framework that jointly addresses regression and classification under uncertainty by combining Perceiver IO~\cite{perceiverio} fusion,a hierarchical Bayesian context layer, evidential NIG regression, SNGP classifier. Our ablation study reveals that combination  of student, instructor, screen provides the lowest error across all metrics.
The Bayesian context layer learns distinguishable per-student posteriors, with a strong negative correlation between clip count and KL divergence from the population prior, confirming that hierarchical regularisation correctly controls personalization as a function of data availability.
On the CASED challenge leaderboard the system achieves competitive regression performance while producing calibrated epistemic uncertainty estimates at inference time. As a future work extending the Bayesian context layer with explicit personality-trait conditioning and cross-session identity tracking are promising directions for further improving personalised engagement modelling.


\section*{Safe and Responsible Innovation Statement}
This work processes video recordings of students in tutoring sessions, raising inherent privacy concerns. All data used in this study was collected under informed consent as part of the CASED~\cite{cased,cased2} challenge, DaiSEE~\cite{daisee} and Aff-Wild2~\cite{abaw1,abaw2,abaw3} datasets. The model relies on face analysis and behavioral signal extraction, which may exhibit performance disparities across demographic groups not well-represented in the relatively small training cohort. The system is intended as a research tool for understanding engagement dynamics, not as a surveillance or performance evaluation instrument for deployed educational settings.

\bibliographystyle{ACM-Reference-Format}
\bibliography{bibfile}

@String{Computing = "Computing" }

@String{Computer = "{IEEE} Computer" }

@String{Springer = "Springer-Verlag" }

@article{fredricks2004school,
  title={School engagement: Potential of the concept, state of the evidence},
  author={Fredricks, Jennifer A and Blumenfeld, Phyllis C and Paris, Alison H},
  journal={Review of educational research},
  volume={74},
  number={1},
  pages={59--109},
  year={2004},
  publisher={Sage Publications Sage CA: Thousand Oaks, CA}
}

@article{vjepa2,
  title={V-jepa 2: Self-supervised video models enable understanding, prediction and planning},
  author={Assran, Mido and Bardes, Adrien and Fan, David and Garrido, Quentin and Howes, Russell and Muckley, Matthew and Rizvi, Ammar and Roberts, Claire and Sinha, Koustuv and Zholus, Artem and others},
  journal={arXiv preprint arXiv:2506.09985},
  year={2025}
}

@article{amini2020deep,
  title={Deep evidential regression},
  author={Amini, Alexander and Schwarting, Wilko and Soleimany, Ava and Rus, Daniela},
  journal={Advances in neural information processing systems},
  volume={33},
  pages={14927--14937},
  year={2020}
}

@dataset{cased,
  author       = {Sharma, Gulshan and
                  Li, Jialin and
                  Salam, Hanan},
  title        = {SMART Challenge Series: Context-Aware Student
                   Engagement Detection
                  },
  month        = mar,
  year         = 2026,
  publisher    = {Zenodo},
  doi          = {10.5281/zenodo.19322996},
  url          = {https://doi.org/10.5281/zenodo.19322996},
}

@article{abaw1, title={ABAW: Learning from Synthetic Data \& Multi-Task Learning Challenges}, author={Kollias,
Dimitrios}, journal={arXiv preprint arXiv:2207.01138}, year={2022} }

@inproceedings{abaw2, title={Abaw: Valence-arousal estimation, expression recognition, action unit detection
\& multi-task learning challenges}, author={Kollias, Dimitrios}, booktitle={Proceedings of the IEEE/CVF Conference on
Computer Vision and Pattern Recognition}, pages={2328--2336}, year={2022} }

@inproceedings{abaw3, title={Analysing affective behavior in the second abaw2
competition}, author={Kollias, Dimitrios and Zafeiriou, Stefanos}, booktitle={Proceedings of the IEEE/CVF International
Conference on Computer Vision}, pages={3652--3660}, year={2021}}

@article{daisee,
  title={Daisee: Towards user engagement recognition in the wild},
  author={Gupta, Abhay and D'Cunha, Arjun and Awasthi, Kamal and Balasubramanian, Vineeth},
  journal={arXiv preprint arXiv:1609.01885},
  year={2016}
}

@inproceedings{kendall2018multi,
  title={Multi-task learning using uncertainty to weigh losses for scene geometry and semantics},
  author={Kendall, Alex and Gal, Yarin and Cipolla, Roberto},
  booktitle={Proceedings of the IEEE conference on computer vision and pattern recognition},
  pages={7482--7491},
  year={2018}
}

@article{sngp,
  title={Simple and principled uncertainty estimation with deterministic deep learning via distance awareness},
  author={Liu, Jeremiah and Lin, Zi and Padhy, Shreyas and Tran, Dustin and Bedrax Weiss, Tania and Lakshminarayanan, Balaji},
  journal={Advances in neural information processing systems},
  volume={33},
  pages={7498--7512},
  year={2020}
}

@article{perceiverio,
  title={Perceiver IO: A General Architecture for Structured Inputs \& Outputs},
  author={Andrew Jaegle and Sebastian Borgeaud and Jean-Baptiste Alayrac and Carl Doersch and Catalin Ionescu and David Ding and Skanda Koppula and Andrew Brock and Evan Shelhamer and Olivier J. H'enaff and Matthew M. Botvinick and Andrew Zisserman and Oriol Vinyals and Jo{\~a}o Carreira},
  journal={ArXiv},
  year={2021},
  volume={abs/2107.14795},
  url={https://api.semanticscholar.org/CorpusID:236635379}
}

@inproceedings{clip,
  title={Learning transferable visual models from natural language supervision},
  author={Radford, Alec and Kim, Jong Wook and Hallacy, Chris and Ramesh, Aditya and Goh, Gabriel and Agarwal, Sandhini and Sastry, Girish and Askell, Amanda and Mishkin, Pamela and Clark, Jack and others},
  booktitle={International conference on machine learning},
  pages={8748--8763},
  year={2021},
  organization={PmLR}
}

@article{audiomae,
  title={Masked autoencoders that listen},
  author={Huang, Po-Yao and Xu, Hu and Li, Juncheng and Baevski, Alexei and Auli, Michael and Galuba, Wojciech and Metze, Florian and Feichtenhofer, Christoph},
  journal={Advances in neural information processing systems},
  volume={35},
  pages={28708--28720},
  year={2022}
}

@inproceedings{kaldiaudio,
  title={The Kaldi speech recognition toolkit},
  author={Povey, Daniel and Ghoshal, Arnab and Boulianne, Gilles and Burget, Lukas and Glembek, Ondrej and Goel, Nagendra and Hannemann, Mirko and Motlicek, Petr and Qian, Yanmin and Schwarz, Petr and others},
  booktitle={IEEE 2011 workshop on automatic speech recognition and understanding},
  year={2011},
  organization={IEEE Signal Processing Society}
}

@inbook{engagement_zimmer,
title = "Student engagement: What is it? Why does it matter?",
author = "Finn, \{Jeremy D.\} and Zimmer, \{Kayla S.\}",
note = "Publisher Copyright: {\textcopyright} Springer Science+Business Media, LLC 2012. All rights reserved.",
year = "2012",
month = jan,
day = "1",
doi = "10.1007/978-1-4614-2018-7\_5",
language = "English",
isbn = "9781461420170",
pages = "97--131",
booktitle = "Handbook of Research on Student Engagement",
publisher = "Springer US",
address = "United States",
}

@inproceedings{DBLP:conf/iclr/LoshchilovH19,
  author       = {Ilya Loshchilov and
                  Frank Hutter},
  title        = {Decoupled Weight Decay Regularization},
  booktitle    = {7th International Conference on Learning Representations, {ICLR} 2019,
                  New Orleans, LA, USA, May 6-9, 2019},
  publisher    = {OpenReview.net},
  year         = {2019},
  url          = {https://openreview.net/forum?id=Bkg6RiCqY7},
  bibsource    = {dblp computer science bibliography, https://dblp.org}
}

@article{bergdahl2024unpacking,
  title={Unpacking student engagement in higher education learning analytics: a systematic review},
  author={Bergdahl, Nina and Bond, Melissa and Sj{\"o}berg, Jeanette and Dougherty, Mark and Oxley, Emily},
  journal={International Journal of Educational Technology in Higher Education},
  volume={21},
  number={1},
  pages={63},
  year={2024},
  publisher={Springer}
}

@inproceedings{annotation,
author = {Khan, Shehroz and Safa, Sadaf},
title = {Revisiting Annotations in Online Student Engagement},
year = {2024},
isbn = {9798400709319},
publisher = {Association for Computing Machinery},
address = {New York, NY, USA},
url = {https://doi.org/10.1145/3641181.3641186},
doi = {10.1145/3641181.3641186},
booktitle = {Proceedings of the 2024 10th International Conference on Computing and Data Engineering},
pages = {111–117},
numpages = {7},
location = {Bangkok, Thailand},
series = {ICCDE '24}
}

@article{screencontent,
  title={Understanding the cognitive cost of multimedia learning: effects of visual load and language proficiency},
  author={Bali, Cintia and Tasdelen, Buket and Bandi, Szabolcs and Zsid{\'o}, Andr{\'a}s},
  journal={Cognitive Research: Principles and Implications},
  volume={11},
  number={1},
  pages={2},
  year={2026},
  publisher={Springer}
}

@article{inconsistency,
  title={Inconsistencies in Measuring Student Engagement in Virtual Learning - A Critical Review},
  author={Shehroz S. Khan and Alireza Abedi and Tracey J. F. Colella},
  journal={ArXiv},
  year={2022},
  volume={abs/2208.04548},
  url={https://api.semanticscholar.org/CorpusID:251442804}
}

@inproceedings{cased2,
author = {Li, Jialin and Sharma, Gulshan and Salam, Hanan},
title = {Personality-Aware Engagement Prediction in Online Learning},
year = {2025},
isbn = {9798400720529},
publisher = {Association for Computing Machinery},
address = {New York, NY, USA},
url = {https://doi.org/10.1145/3746270.3760234},
doi = {10.1145/3746270.3760234},
booktitle = {Proceedings of the 3rd International Workshop on Multimodal and Responsible Affective Computing},
pages = {119–127},
numpages = {9},
location = {Ireland},
series = {MRAC '25}
}

@ArtifactSoftware{R,
    title = {R: A Language and Environment for Statistical Computing},
    author = {{R Core Team}},
    organization = {R Foundation for Statistical Computing},
    address = {Vienna, Austria},
    year = {2019},
    url = {https://www.R-project.org/},
}

\end{document}